\documentclass{article}
\usepackage[square]{natbib}
\setcitestyle{numbers,square,comma}
\usepackage[utf8]{inputenc} % allow utf-8 input
\usepackage[T1]{fontenc}    % use 8-bit T1 fonts
\usepackage{hyperref}       % hyperlinks
\usepackage{url}            % simple URL typesetting
\usepackage{booktabs}       % professional-quality tables
\usepackage{amsfonts}       % blackboard math symbols
\usepackage{nicefrac}       % compact symbols for 1/2, etc.
\usepackage{microtype}      % microtypography
\usepackage{xcolor}         % colors
\usepackage{graphicx}
\usepackage{wrapfig}
\usepackage{cleveref}
\usepackage[preprint]{neurips_2024}

\title{TIGER: Text-Instructed 3D Gaussian Retrieval and Coherent Editing}

\author{%
  Teng Xu~~~~Jiamin Chen~~~~Peng Chen~~~~Youjia Zhang~~~~Junqing Yu~~~~Wei Yang\footnotemark[2]\\
  Huazhong University of Science and Technology\\
  \texttt{\{tengxu, jiaminchen, pengchen, youjiazhang, yjqing, weiyangcs\}@hust.edu.cn} \\
}

\begin{document}

\maketitle
\renewcommand{\thefootnote}{\fnsymbol{footnote}}
\footnotetext[2]{Corresponding Author}

\begin{abstract}

%Editing objects within a scene, with recent research interest of represented as discrete 3D Gaussians,
%is a critical functionality required across a broad spectrum of applications.
Editing objects within a scene is a critical functionality required across a broad spectrum of applications in computer vision and graphics. 
As 3D Gaussian Splatting (3DGS) emerges as a frontier in scene representation, the effective modification of 3D Gaussian scenes has become increasingly vital.
This process entails accurately retrieve the target objects and subsequently performing modifications based on instructions. 
Though available in pieces, existing techniques mainly embed sparse semantics into Gaussians for retrieval, and rely on an iterative dataset update paradigm for editing, leading to over-smoothing or inconsistency issues.
To this end, this paper proposes a systematic approach, namely TIGER, for coherent text-instructed 3D Gaussian retrieval and editing. In contrast to the top-down language grounding approach for 3D Gaussians, we adopt a bottom-up language aggregation strategy to generate a denser language embedded 3D Gaussians that supports open-vocabulary retrieval. To overcome the over-smoothing and inconsistency issues in editing, we propose a Coherent Score Distillation (CSD) that aggregates a 2D image editing diffusion model and a multi-view diffusion model for score distillation, producing multi-view consistent editing with much  finer details. In various experiments, we demonstrate that our TIGER is able to accomplish more consistent and realistic edits than prior work. 
Result videos can be found on the project website: \href{https://xutanxing.github.io/TIGER/}{https://xutanxing.github.io/TIGER/}.

\end{abstract}

\section{Introduction}
\label{sec:intro}

Object editing within three-dimensional scenes constitutes a critical functionality in digital modeling, with broad applications spanning movie/game development, architecture, virtual reality and etc.
This editing process entails retrieve the objects to be edited and subsequently performing modifications based on instructions.
Recent advancements in neural scene representations have expedited the digitization of real-world 3D scenes from multi-view images~\cite{yu2021plenoctrees, fridovich2022plenoxels, muller2022instant,chen2023mobilenerf}, highlighting the necessity for retrieval and editing techniques that are compatible with these advancements.
Pioneering researches have focused on developing editing methods tailored to implicit 3D scene representations~\cite{park2019deepsdf, liu2020dist, vicini2022differentiable, mescheder2019occupancy, chen2019learning, lombardi2019neural}, particularly the Neural Radiance Field (NeRF)~\cite{mildenhall2021nerf, wang2022clip, park2023ed, wang2023nerf, zhuang2023dreameditor, huang2022stylizednerf}.
Notably, the Instruct-NeRF2NeRF~\cite{haque2023instruct} introduces iterative dataset updating approach that leveraging an pre-trained image editing diffusion model~\cite{brooks2023instructpix2pix} for text-driven NeRF scene editing. However, the implicit nature of NeRF poses significant challenges when instructed to edit a specific target object within the scene.

\begin{figure}[htb]
  \begin{minipage}[t]{1\linewidth} % 将图像包裹在 minipage 环境中，使其宽度为 0.6 行宽
    \includegraphics[width=\linewidth]{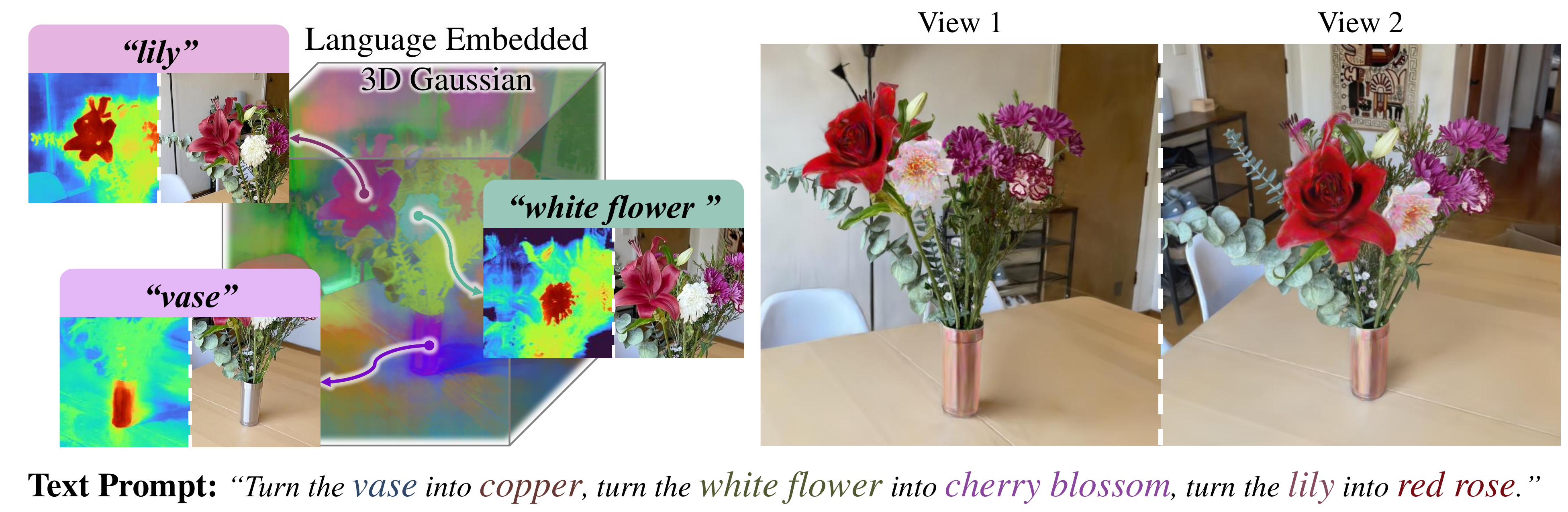}
    \caption{Our TIGER presents a systematic framework for 3D Gausssian retrieval and editing. TIGER integrates language features into each Gaussian primitive, and support open-vocabulary query directly in space. TIGER demonstrates excellent zero-shot retrieval capabilities, and enable detail preserving and multi-view consistent editing.}
    \label{fig:teaser}
    \vspace{-5mm}
  \end{minipage}
\end{figure}
As 3D Gaussian Splatting (3DGS)~\cite{kerbl20233d} emerges as the new frontier in scene representation, the effective modification of 3D Gaussian scenes has attracted significant attention. Unlike implicit methods, 3DGS adopts 3D Gaussians as explicit geometric primitives, providing foundations for retrieval and editing operations. Recent endeavors, such as GaussianEditor-NTU~\cite{chen2023gaussianeditor} and GaussianEditor-HW~\cite{fang2023gaussianeditor}, have adopted analogous methodologies for editing 3D Gaussian scenes. These methods identify the target objects from 2D segmentation of images, then unproject the segmentation masks into space to determine the relevant 3D Gaussians for editing. Subsequently, they use an iterative dataset updating scheme, akin to Instruct-NeRF2NeRF, to optimize the targeted Gaussians based on evolving image inputs. Despite the notable achievements of these approaches, their reliance on image-guided retrieval  necessitates repeated 2D segmentation for each editing operation, relies on optimal rendering perspectives for precise segmentation of the target, and the dataset updating scheme is prone to inducing over-smoothing and the multi-face Janus problem~\cite{haque2023instruct}.

To tackle the aforementioned challenges, this paper introduces a systematic approach termed TIGER (\textbf{T}ext-\textbf{I}nstructed \textbf{G}aussian \textbf{R}etrieval and \textbf{E}diting). TIGER offers direct Gaussian retrieval in 3D space according to textual input, obviating the need for 2D image bridging, and introduces a detail-preserving, multi-view consistent editing paradigm.
Specifically, to enable the retrieval of 3D Gaussians directly from textual prompts, we incorporate language embedding into each 3D Gaussian via differentiable rendering. Previous approaches for Gaussian grounding typically rely on 2D supervision from semantics~\cite{chen2023gaussianeditor, fang2023gaussianeditor} or extract language feature within each distinct mask~\cite{qin2023langsplat, zuo2024fmgs}.
However, such top-down strategies (segment and extract feature within a mask) constrain the diversity of queryable objects as language features are extracted within mask, and hence ignoring context outside mask. In response, we devise a bottom-up scheme utilizing MaskCLIP~\cite{zhou2022extract} to extract nuanced, low-level language features and subsequently employ FeatUp~\cite{fu2024featup} to refine these features into a high-resolution language feature map for 3D Gaussian supervision. Our scheme provides open-vocabulary query directly in 3D space. For retrieval, we measure the relevancy score between each 3D Gaussian's language embedding and the object text query.
Furthermore, for the editing process, we propose a novel Coherent Score Distillation Sampling (CSD) technique that integrates the SDS losses from both an image editing diffusion model and a multi-view diffusion model. We update the 3D Gaussians according to the CSD loss, applying a higher update rate to the more relevant Gaussians, and a lower rate conversely. This technique ensures that edits not only adhere to the textual prompt through the image editing diffusion model but also maintain consistency across multiple views via the multi-view diffusion model. In our implementation, we employ the pre-trained InstructPix2Pix~\cite{brooks2023instructpix2pix} as the image editing diffusion model, which is adept at following detailed text directives for image transformations. We utilize MVDream~\cite{shi2023MVDream} as the multi-view diffusion model to guarantee that the edits remain coherent when observed from different perspectives. This integrated approach facilitates a robust editing framework that upholds both fidelity to textual instructions and spatial consistency.
Extensive experiments demonstrate that our TIGER supports accurate open-vocabulary retrieval, and is able to accomplish more consistent and realistic edits than prior work.

\section{Related Works}
\label{sec:rw}
\paragraph{3D langugae fields.}
The interaction between language and 3D has long been a focal point for vision researchers in the realm of visual studies~\cite{zhi2021place,siddiqui2023panoptic,kobayashi2022decomposing,tschernezki2022neural,gordon2018iqa,azuma2022scanqa,cascante2022simvqa,corona2022voxel,thomason2022language}.
NeRFs~\cite{mildenhall2021nerf}, widely recognized for producing photorealistic new perspectives of a scene from calibrated photographs, have gained popularity and have seen numerous extensions. In the realm of 3D technology, significant advancements have been made in integrating language, revolutionizing our interaction with digital environments. LERF~\cite{kerr2023lerf} embeds CLIP~\cite{radford2021learning} features into NeRF~\cite{mildenhall2021nerf}, enabling more intuitive natural language interaction and providing extensive scene analysis. 3D-OVS~\cite{liu2023weakly} leverages pre-trained foundational models in a weakly supervised manner to distill neural radiance fields, effectively elevating 2D features to view-consistent 3D segmentation. Recent work has also introduced 3D Gaussian representations~\cite{kerbl20233d} into the domain of 3D language localization and segmentation, achieving more precise localization. LangSplat~\cite{qin2023langsplat} proposes using SAM~\cite{kirillov2023segment} to learn hierarchical semantics, eliminating the need for DINO~\cite{caron2021emerging} feature regularization. Additionally, they trained a scene-specific language autoencoder to reduce memory requirements, though their features remain hierarchical, necessitating multiple queries across different levels. FMGS~\cite{zuo2024fmgs} integrates visual language embeddings from foundational models into 3D Gaussian representations. They scale up the embeddings of the smaller scales in the pre-computed CLIP feature pyramid bilinearly to the largest scale feature map and generate a composite feature map by averaging these embeddings. 
In previous work, images were cropped and fed into CLIP to obtain features for each patch. This is also the primary reason why they require hierarchical processing. We, however, employ MaskCLIP~\cite{zhou2022extract} to directly generate dense semantic features, and then use FeatUp~\cite{fu2024featup} to upsample these features to the pixel level. Consequently, we only need to train a single-layer language field, while still maintaining access to global information.

\paragraph{3D editing.}
Editing neural radiance fields (NeRF~\cite{mildenhall2021nerf}) has become a popular research direction recently. However, due to the implicit representation of NeRF, there is a lack of precise localization for the editing objects. As a result, most of the editing work focuses on the entire 3D scene~\cite{huang2021learning,huang2022stylizednerf,nguyen2022snerf,wang2022clip,wu2022palettenerf,zhang2022arf,wang2023nerf}. Some studies have focused on object-centric editing problems. For example, Instruct-NeRF2NeRF~\cite{haque2023instruct} uses text to control 3D scene editing, allowing constraints on the editing object through text. However, it relies solely on InstructPix2Pix~\cite{brooks2023instructpix2pix} to control the background, often resulting in global modifications. DreamEditor~\cite{zhuang2023dreameditor} represents scenes as mesh-based neural fields and accurately edits specified areas based on text prompts. However, the quality and speed of editing are constrained by the scene representation. Recent work has introduced 3D Gaussians~\cite{kerbl20233d} into the field of 3D editing~\cite{fang2023gaussianeditor,chen2023gaussianeditor}, whose explicit representation can accurately localize the editing areas. GaussianEditor-NTU~\cite{chen2023gaussianeditor} introduces Gaussian semantic tracing, achieving precise editing. However, it still relies on an iterative dataset update paradigm for editing, which leads to issues such as over-smoothing or inconsistency. We successfully apply Coherent Score Distillation to 3D Gaussian editing, producing results that better adhere to the editing instructions and exhibit richer details.

\section{Method}
\label{sec:lg}
In this section, we introduce our TIGER method that offers open-vocabulary 3D Gaussian retrieval and detail-preserving, multi-view consistent editing.
Specifically, we incorporate language embedding into each 3D Gaussian with supervision from a bottom-up language feature extraction scheme (\cref{sec:lang_gaussian}). And we propose a novel Coherent Score Distillation Sampling (CSD) technique that integrates the SDS losses from both an image editing diffusion model and a multi-view diffusion model (\cref{sec:edit}) for consistent and realistic editing.

\subsection{Language Embedded 3D Gaussian for Open-vocabulary Retrieval}
\label{sec:lang_gaussian}

Grounding language embedding into 3D Gaussians entails adding a language feature into each Gaussian primitive and optimizing them through supervision from images.
%
 
%
%, typically the CLIP features~\cite{xxx}, into 3D Gaussians . However, .
%, as recent research advancements have demonstrated, signifies a leap forward in scene understanding. 
%Our approach builds upon these methods, which typically require extracting multi-resolution features to locate objects at different scales, given that CLIP features~\cite{radford2021learning} are at the image level. 
\begin{figure}[htb]
  \includegraphics[width=1\linewidth]{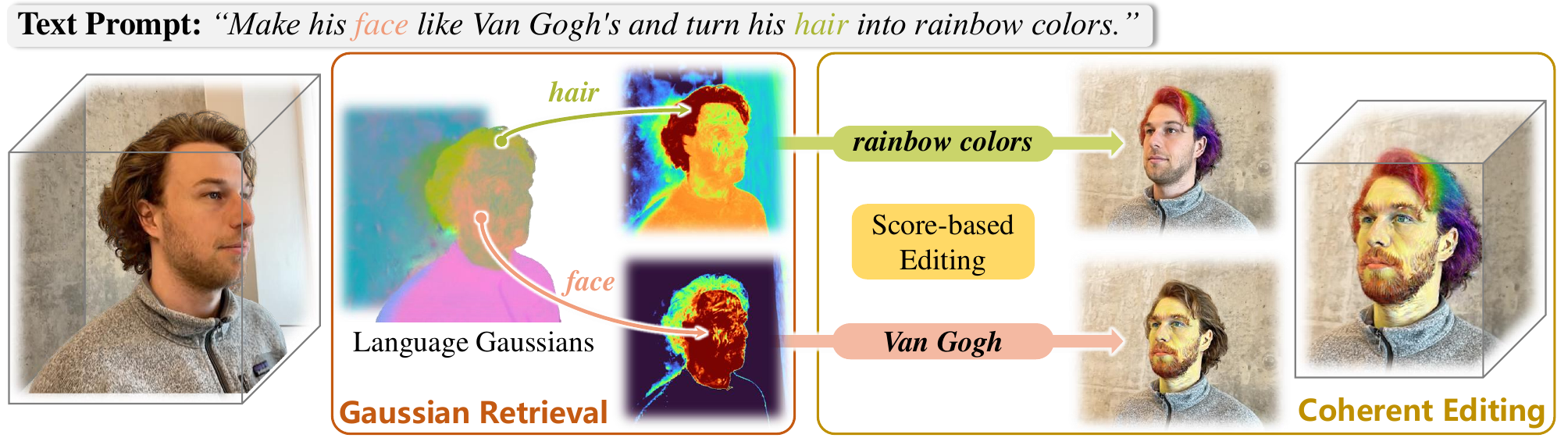}
  \caption{The pipeline of our method. We first embed language features into each Gaussian primitive. Upon receiving editing prompt, we compute a relevance score for each Gaussian w.r.t. the given edit prompt. Subsequently, we can update Gaussians using our CSD based on the relevancy scores.}
  \label{fig:pipeline}
\end{figure}
\begin{figure}[htb]
  \begin{minipage}[t]{1\linewidth} % 将图像包裹在 minipage 环境中，使其宽度为 0.6 行宽
    \includegraphics[width=\linewidth]{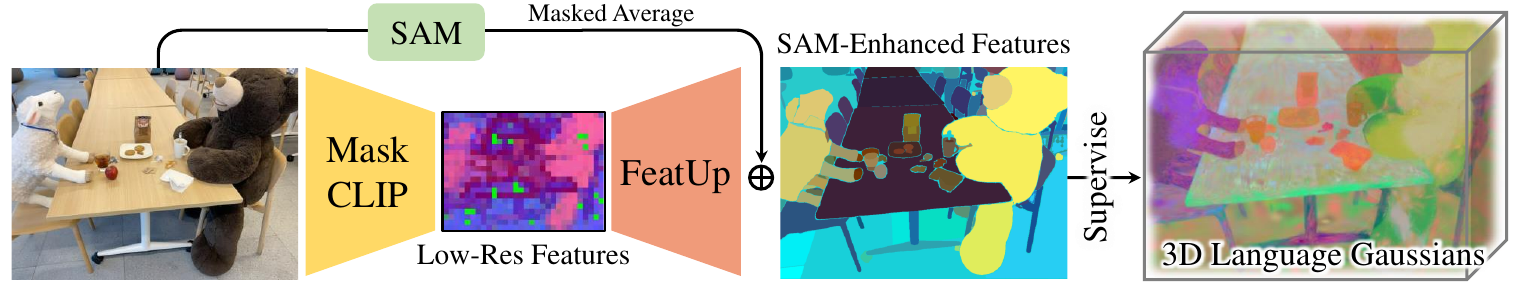}
    \caption{Our Language Embedding Process: we use MaskCLIP to generate low-resolution semantic features with global context information, then upsample the low-resolution features into high-resolution for 3D Gaussian language supervision using FeatUp~\cite{fu2024featup}. To better preserve the sharp boundery, we apply SAM to the finest level and aggregate features with each fine mask. Finally, the refined language features are embedded into 3D Gaussians via differentiable rendering, enabling precise retrieval of relevant Gaussian points based on open-vocabulary query.}
    \label{fig:local_pipeline}
  \end{minipage}
\end{figure}

\paragraph{3D Language Gaussians}
3D Gaussian Splatting represents a scene as a volume of anisotropic Gaussians, where each Gaussian is characterized by its position $u \in \mathbb{R}^3 $, covariance $\Sigma \in \mathbb{R}^{7}$, color $c  \in \mathbb{R}^3 $ and opacity $o \in \mathbb{R}$. We specifically add another element, the language embedding vector $\mathcal{L} \in \mathbb{R}^{64} $ into each Gaussian and form the primitive of our representation: $\mathcal{G} = \{ u, \Sigma, c, o, \mathcal{L} \}$.
%\begin{equation}
%  \mathcal{G} = \{ u, \Sigma, c, \alpha, L \}
%  \label{eq:3d_primitive}
%\end{equation}
% 写3D Gaussian怎么render出color和embedding
%For a point $p$ in physical 3D space, each Gaussian influences its color according to the standard Gaussian equation, as: $f_c (p) = \exp \big ( -\frac{1}{2} (p - u)^{T} \Sigma^{-1} (p - u) \big )$.
%\begin{equation}
%  f_c (p) = \exp \big ( -\frac{1}{2} (p - u)^{T} \Sigma^{-1} (p - u) \big )
%  \label{eq:color}
%\end{equation}
Blend ordered Gaussians $\mathcal{G}_i, i=1...N$ along a pixel ray $r$ and denote $\alpha_i = o_i G(x_i)$ which represents the influence of the $i$-th Gaussian to the image pixel, %we can obtain the color of the rendered pixel.
%\begin{equation}
%  C = \sum_{i \in N }f_c^i{\alpha}_i\prod_{j=1}^{i-1} (1-{\alpha}_j)
%  \label{eq:render_color}
%\end{equation}
%The covariance matrix $\Sigma$ can be further decomposed into a scaling matrix and a rotation matrix, then deformed by transformations pertaining to the camera viewpoint, for differentiable optimization from color images.
%The language embedding vector $L$ is optimized with the objective: SAM-Enhanced Features $F_s$. With the language-embedded 3D Gaussians, 
we can render out the language feature $f$ of the pixel via a blending process:
%\begin{equation}
%  f = \sum_{{G}_i (L) \in r} \mathcal{G}_i (L) {\alpha}_i\prod_{j=1}^{i-1} (1-{\alpha}_j)
%  \label{eq:important}
%\end{equation}
\begin{equation}
  f = \sum_{i = 1}^{N} (\mathcal{L}_i {\alpha}_i\prod_{j=1}^{i-1} (1-{\alpha}_j))
  \label{eq:important}
\end{equation}

\paragraph{Bottom-up Language Feature Extraction}
% We obtain the low-resolution CLIP features of the images from training views through MaskCLIP. LanSplat has three different semantic levels and generates three different masks for a point prompt using SAM: whole, part, and subpart. Each 3D Gaussian has three different levels of language embeddings and these embeddings are obtained from CLIP features of the masked image patches. In contrast, our method requires only one semantic level, hence only one language field needs to be trained. With MaskCLIP serving as the backbone, these features are then upsampled to the original image resolution using the FeatUp module. We apply the smallest scale SAM from LanSplat for global segmentation to generate a global mask. We then average the upsampled CLIP features for each mask to obtain the SAM-Enhanced Features. Finally, we use the SAM-Enhanced Features as supervise to train a 3D language field. This significantly reduces training time and resource consumption. Besides, LanSplat trains an autoencoder to compress the CLIP features into 3 dimensions for each scene, which leads to ???. We leverage PCA to reduce the SAM-Enhanced Features from 512 dimensions to 64 dimensions and use the SAM-Enhanced Features as supervision to train the language field. This approach ???.
To optimize the language embedding within each Gaussian, it is crucial to first transform the 2D image dataset into a per-pixel language feature map for effective supervision using Eqn.~\ref{eq:important}. The fidelity and granularity of this feature map are critical, as they directly influence the quality of the final language embedding. However, the language feature, i.e., CLIP~\cite{radford2021learning}, is extract from the whole image. 
Exiting endeavors, such as GaussianEditor-NTU~\cite{chen2023gaussianeditor}, GaussianEditor-HW~\cite{fang2023gaussianeditor}, rely on masks to tag each Gaussian, hence rely on 2D images for retrieving. LangSplat~\cite{qin2023langsplat} and Gaussian Grouping~\cite{ye2023gaussian}, segment the images first and extract CLIP features~\cite{radford2021learning} within each mask. However, these top-down strategies (segment and then extract language feature within each mash) constrains the capability for open-vocabulary queries. 
We propose an innovative bottom-up extraction method enabling precise and open-vocabulary retrieval of objects, laying a solid foundation for subsequent 3D editing tasks.

%Exiting top-down approaches (segment and extract language features within each mask), we introduce a bottom-up language feature extraction approach.
%To optimize the language embedding in each Gaussian, it is first necessary to convert the 2D image dataset into a language feature map for supervision. The quality of this feature map largely determines the quality of the Language Embedded Representation. We propose a method that can achieve high-resolution, numerically uniform, and edge-clear feature maps. An overview is provided in Fig.~\ref{fig:local_pipeline}.

To generate the language feature map, we first use MaskCLIP to produce low-resolution patch-level language features. This patch feature  encompass multi-scale global information through masked self-distillation features. This is fundamentally different from extracting features by cropping the image first and then inputting it into Clip.  
We then use FeatUp to upsample these features to pixel-level language features. To further save memory, we use PCA to the extracted CLIP feature dimension to 64 before supervision. During language embedding optimization, we only update $\mathcal{L}$ of each Gaussian while keeping all other properties unchanged. 

%Although using neural networks for compression, as in LangSplat, would save more memory, the network tends to overfit the training data. If the optimized results shift, it would not be able to decompress effectively.
%
Although the language feature is sufficient resolution for Gaussian supervision at this stage, they still suffer uneven bounderies.%, and the semantic information of objects tends to bleed into the background. 
For further refinement, we apply SAM~\cite{kirillov2023segment} at the finest level, and generate a set of fine binary masks ${M}$, we then conduct masked average to aggregate all the language features within each mask to obtain refined semantic boundaries. Notice our masked average process is fundamentally different to the top-down approach as our feature generated from MaskCLIP contains global information that across semantic boundaries. While extracting language features within each mask lead of the absence of context information outside the mask in the final language feature.

%, inspired by LangSplat, we introduce SAM to enhance the semantic features.
%
%We first use SAM to generate a set of fine binary masks ${M}$, where a value of 1 in $M_i$ indicates the pixels included in the $i$-th mask. Let $F$ represent the pixel-level semantic features generated by FeatUp. We use Masked Average to aggregate all the semantic features within each mask.
%\begin{equation}
%{F}_{i}=\frac{\sum \mathbf{M}_{i} \cdot \mathbf{F}}{\sum \mathbf{M}_{i}}
%\end{equation}
%Then, we assign $F_i$ to all the pixels in $M_i$  to obtain the SAM-Enhanced feature map $F_s$, As shown in Fig.~\ref{fig:local_pipeline}. The visualization of the features is achieved by compressing the features using PCA. To save memory, we also use PCA to compress $F_s$ before supervision. Although using neural networks for compression, as in LangSplat, would save more memory, the network tends to overfit the training data. If the optimized results shift, it would not be able to decompress effectively.

\paragraph{Open-vocabulary 3D Gaussian Retrieval}
After optimization, we can directly retrieve 3D Gaussians using similarity between $\mathcal{L}$ and the textual object query $\mathcal{T}$. Thanks to the bottom-up feature extraction, our language embedding contains both local and global context information, hence support open-vocabulary querying.
%Since the CLIP model encodes text and images into a unified latent space, our learned 3D language Gaussians can achieve open-vocabulary querying. 
Given an arbitrary object query, our method achieves excellent zero-shot localization capability, as shown in Fig.~\ref{fig:loc_compare}. It's quite intuitive to directly retrieve an object $\mathcal{O}$ using a embedded text prompt $\mathcal{T}^*$ according to the cosine similarity with their language embedding $\mathcal{L}$.
% 写3D Gaussian怎么render出color和embedding
\begin{equation}
  \mathcal{O} = \{\mathcal{G}: \langle \mathcal{T}^* \cdot \mathcal{G} (\mathcal{L}) \rangle > \tau \}
  \label{eq:render_relevancy}
\end{equation}
where $\tau$ is the relevancy threshold. 
%For the image level, we compute the relevancy score for each text query following LERF. 
Notice our representation also supports image level query as we can render the language embedding into images according to Eq.~\ref{eq:important}, and calculate the relevancy score. 
Different from LERF and LanSplat, which require computing relevancy score across multiple scales and levels and taking the highest score, our method only needs to conduct one level of computation.
\subsection{Coherent Score Distillation for Gaussian Editing}
\label{sec:edit}
% \begin{figure}[htb]
%   \begin{minipage}[t]{0.7\linewidth} % 将图像包裹在 minipage 环境中，使其宽度为 0.6 行宽
%     \includegraphics[width=\linewidth]{fig/style_pipe1.pdf}
%     \caption{The pipeline of our approach.}
%     \label{fig:style_pipeline}
%   \end{minipage}
% \end{figure}

\begin{wrapfigure}{r}{0.7\linewidth} % 'r' 表示图像靠右，宽度为 0.7 行宽
  \centering
  \vspace{-4mm}
  \includegraphics[width=\linewidth]{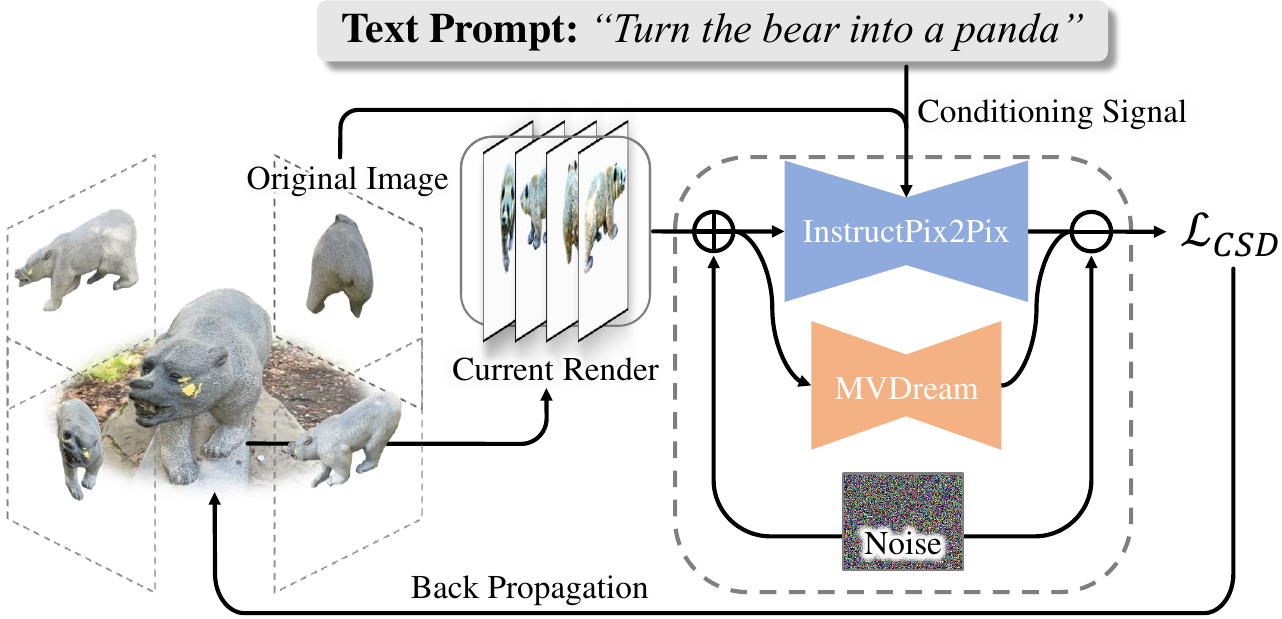}
  \caption{Our 2D Gaussian editing method use a Coherent Score Distillation that leverages 2D image editing diffusion model (InstructPix2Pix) for instruct-based editing and utilizes multi-view diffusion model (MVDream) to address multi-face inconsistency issue, and achieve multi-view consistent edits with fine details.}
  \label{fig:style_pipeline}
\end{wrapfigure}

Building upon the language-embedded 3D Gaussians, we can perform scene editing instructed by the text prompts. Our method takes the reconstructed 3D Gaussian scene and language prompt $\mathcal{T}$ for editing as input. As output, our method generates an edited version of the Gaussian scene according to the provided instructions. 
Existing 3D Gaussian editing methods unanimously adopt the iterative dataset updating scheme~\cite{chen2023gaussianeditor, fang2023gaussianeditor}, which induces over-smoothing and multi-face Janus problem as the editing of each image is independent.
To address the issue, we propose a novel Coherent Score Distillation (CSD) that integrate the SDS losses of a 2D image editing diffusion model, i.e., the InstructPix2Pix, and a multi-view diffusion model, i.e., the MVDream, producing multi-view consistent editing with fine details. An overview is provided in Fig.~\ref{fig:style_pipeline}.

For a 3D Gaussian scene $ \mathbf{S} = \{ \mathcal{G}_i \} $, we retrieve 3D Gaussians according to the relevance scores w.r.t. the text prompt, and calculate the center and bounding box of the targeted object. Accordingly, we then choose four camera views ${v_i, i=1,2,3,4}$ around the targeted object (every $90^\circ$  for $360^\circ$ scenes, and evenly distributed between bounds for other scenarios) and render the corresponding views $\mathbf{x}_{v_i} = \mathcal{R}(\mathbf{S}, v_i)$, where $\mathcal{R}$ denotes the differentiable rendering function. We use $\epsilon_{\phi}$ to denote the InstructPix2Pix model, which take a noisy image $\mathbf{x}_{v_i}^t = \mathbf{x}_{v_i} + \sigma_t \epsilon$ and edit text $\mathcal{T}$ as input and output the noise to be reduced.
Similarly, we use $\epsilon_{\psi}$ to denote the MVDream model, which take noisy images from all four views $\{\mathbf{x}_{v_i}^t\}$, text $\mathcal{T}$ and camera view $v_i$ as input and output the noise to be reduced of all four views.
Consequently, the averaged noise residuals of $\epsilon_{\phi}$ and $\epsilon_{\psi}$ on rendered views are as follows:
\begin{equation}
%\begin{aligned}
\mathcal{L}_{\mathbf{ip}} =\mathbb{E}_{t, \epsilon, i} \left[ \omega(t)\left(\epsilon_{\phi}\left(\mathbf{x}_{v_i}^t; t, \mathbf{x}_{v_i}, \mathcal{T}\right)-\epsilon\right) \right]
,
%\end{equation}
%
%\begin{equation}
\mathcal{L}_{\mathbf{mv}}  =\mathbb{E}_{t, \epsilon}\left[\omega(t)\left(\epsilon_{\psi}\left( 
\{\mathbf{x}_{v_i}^t\}; t, \{ v_i \}, \mathcal{T}\right)-\epsilon\right) \right]
%\end{aligned}
\end{equation}
%
%
%\begin{equation}
%\nabla_{\mathbf{S}} \mathcal{L}_{\mathbf{CSD}}=\mathbb{E}_{t, \epsilon}\left[\omega(t_1)\left(\epsilon_{\phi}\left(\mathbf{x}_{v_i}^t; t, \mathbf{x}_{v_i}, \mathcal{T}\right)-\epsilon\right) + \omega(t)\left(\epsilon_{\phi}\left(\mathbf{x}_{v_i}^t; t, \mathbf{x}_{v_i}, \mathcal{T}\right)-\epsilon\right) \frac{\partial \mathbf{x}_{v_i}}{\partial \mathbf{S}}\right]
%\end{equation}
%
%
%\begin{equation}
%\begin{aligned}
%\nabla_{\mathbf{S}}\mathcal{L}_{\mathbf{ip}} (\mathbf{S}) =\mathbb{E}_{t, \epsilon} \left[ \omega(t)\left(\epsilon_{\phi}\left(\mathbf{x}_{v_i}^t; t, \mathbf{x}_{v_i}, \mathcal{T}\right)-\epsilon\right) \frac{\partial \mathbf{x}_{v_i}}{\partial \mathbf{S}}\right]
%,\,\,\,
%\end{equation}
%
%\begin{equation}
%\nabla_{\mathbf{S}} \mathcal{L}_{\mathbf{mv}} (\mathbf{S}) =\mathbb{E}_{t, \epsilon}\left[\omega(t)\left(\epsilon_{\psi}\left(\mathbf{x}_{v_i}^t; t, \mathbf{x}_{v_i}, v_i\right)-\epsilon\right) \frac{\partial \mathbf{x}_{v_i}}{\partial \mathbf{S}}\right]
%\end{aligned}
%\end{equation}
%
where $\omega(t)$ is a weighting function. Finally, we have our CSD as the combination of two SDS as:
\begin{equation}
\nabla_{\mathbf{S}} \mathcal{L}_{\mathbf{CSD}} (\mathbf{S}) =\mathbb{E}_{v_i} \left[ (\lambda_1 \mathcal{L}_{\mathbf{ip}} + \lambda_2 \mathcal{L}_{\mathbf{mv}}) \frac{\partial \mathbf{x}_{v_i}}{\partial \mathbf{S}} \right]
%\end{aligned}
\end{equation}
%
%First, we compute the relevance scores of all Gaussians based on the prompt. Using these scores, we calculate the center and radius of the rendering viewpoint, with the rendered image serving as the original image $I$. 
$\lambda_1$ and $\lambda_2$ are weighting factors. For each iteration, we update the 3D Gaussians $\mathbf{S}$ using CSD loss, and then random a new set of four views, and add noises for the next round CSD loss calculation. 
%To be more specific, for a Gaussian scene, we choose four views and render corresponding images, we then add random noises into each image and calculate the differences between predicted noises and added noises using InstructPix2Pix and MVDream. We combine the noise residuals and compute the gradients for Gaussian optimization. 
In practice, we only add minor noise ($t$ is chosen from 0.02 to 0.2) at each optimization step to ensure stable updates to the scene. And during score distillation, we always use the original 3D Gaussian rendered views as conditions for InstructPix2Pix,
%our diffusion model always uses the original rendered images as the image condition,
thus effectively preventing unstable updates due to the excessive fluidity of 3D Gaussians as pointed out by GaussianEditor-NTU. Additionally, we employ a dynamic weighting strategy to blend the gradients of these two diffusion models, initially emphasizing the role of the multi-view diffusion model to prioritize generating consistent geometric structures. Later, we gradually increase the weight of the image editing diffusion model to sculpt details. 
\paragraph{Score-based Updating}
Each of our 3D Gaussian primitive  is associated with a language embedding vector $\mathcal{L}$. With this language attribute, we compute its relevance score with respect to a textual query embedding $\mathcal{T}^*$ using cosine similarity, defined as $s_i = \langle \mathcal{T}^* \cdot \mathcal{G}_i (\mathcal{L}) \rangle$. To optimize the gradient of each Gaussian, we leverage the relevance score $s_i$ to adjust the updating rate. This strategy enables higher update rates for more relevant Gaussians and smaller, or even zero, updates for less relevant ones, facilitating precise object-level editing. For 3D Gaussian updating and densification process, newly added Gaussians inherit the language vector $\mathcal{L}$ attribute of their parent Gaussians, and we selectively densify only the 3D Gaussians within the top $1\%$ of gradient for each iteration to ensure more stable updates.

%We constrain the gradient of each Gaussian using its score $s_i$ to  apply higher updating rate to more relevant Gaussians and smaller or even zero updates to less relevant ones, thereby achieving precise object-level editing. We can also apply an additional threshold to only render the targeted object, accelerating the editing process. But this is only validate when the whole object is targeted as the editing process requires other image parts of reference. 

\section{Experiments}
\label{sec:exp}

In this section, we conduct extensive experiments to demonstrate the effectivenss of our TIGER. %framework.

\subsection{Comparison with SOTA}

\begin{wraptable}{r}{0.5\linewidth}
    \vspace{-10mm}
  \centering
  
  \vspace{-\intextsep} % Adjust the space above the table
  \setlength{\tabcolsep}{1mm}{}
  \begin{tabular}{@{}lcccc@{}}
    \toprule
    \textbf{Test Scene} & \textbf{LERF} & \textbf{LangSplat} & \textbf{Ours} \\
    \midrule
    figurines & 75.0\% & 75.5\% & \textbf{83.7\%}\\
    
    teatime & 84.8\% & \textbf{91.5\%} & 84.8\%\\
    
    ramen & 62.0\% & 67.7\% & \textbf{91.9\%}\\

    waldo kitchen & 72.7\% & 75.0\% & \textbf{87.5}\%\\ 
    \midrule
    Overall & 73.6\% & 77.4\% & \textbf{87.0}\%\\
    \bottomrule
  \end{tabular}
  % \vspace{2mm} % Adjust the space between the table and the caption
  \caption{Localization accuracy comparison. Overall is calculated as the average across scenes.}
  \label{tab:Localization}
\end{wraptable}

\textbf{Language Instructed Localization}

\begin{minipage}{1\linewidth}
To evaluate the open-vocabulary retrieval performance of our TIGER method, we compare with LERF and LangSplat which support open-vocabulary queries. We assess the 3D localization performance of our method using the LERF dataset~\cite{kerr2023lerf}. The LERF dataset contains several large-scale scenes for object localization and retrieval. Following LERF, we use localization accuracy as the evaluation metric. We utilize the test views defined in LangSplat and employ text queries from both LERF and LangSplat for evaluation. 
\end{minipage}

\begin{wrapfigure}{r}{1.05\linewidth} % 'r' 表示图像靠右，宽度为 0.7 行宽
  \centering
  \vspace{-55mm}
  \includegraphics[width=\linewidth]{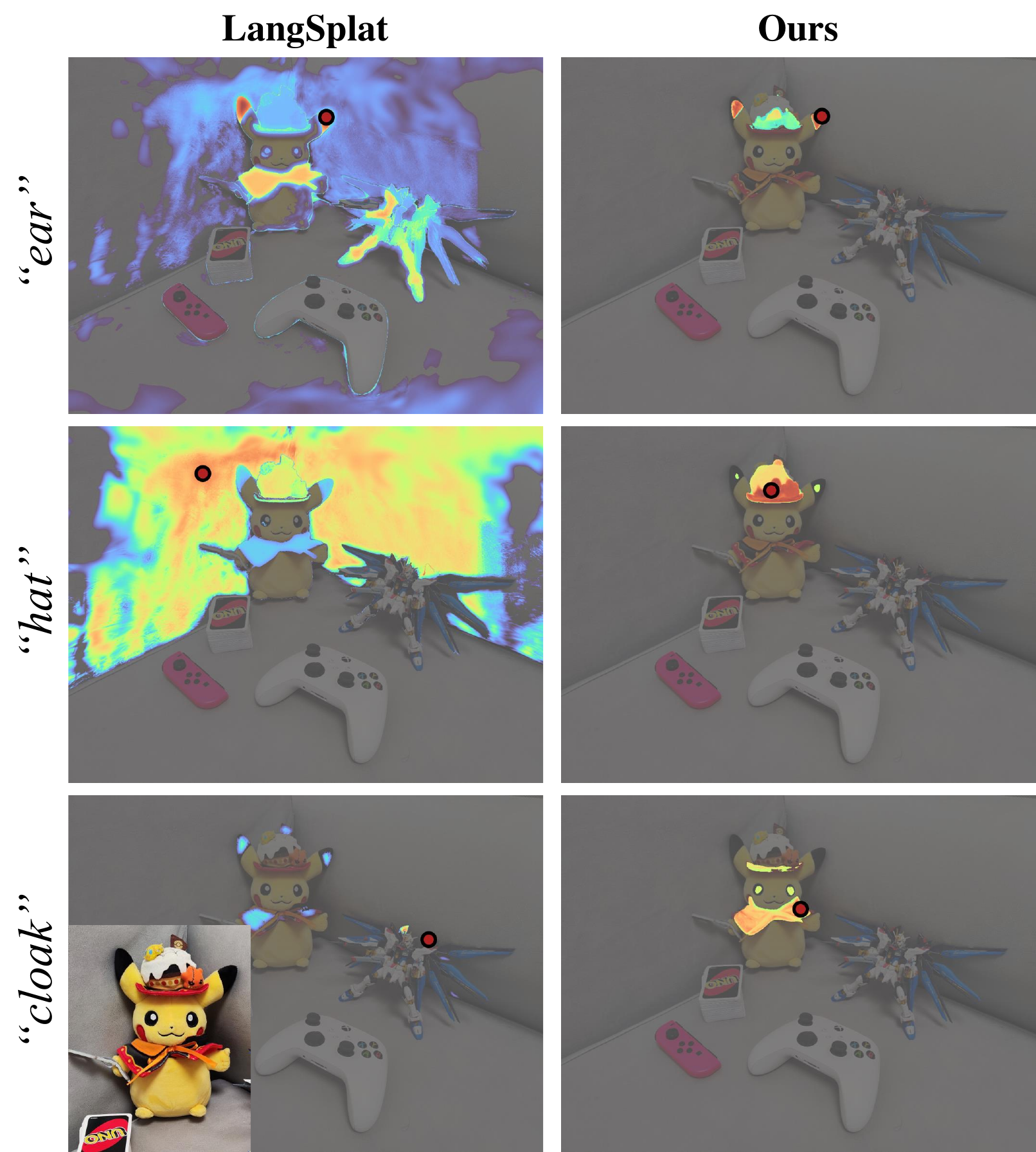}
  \caption{Qualitative comparison: TIGER method performs well in fine-grained localization.}
  \label{fig:sofa_com}
\end{wrapfigure}

As shown in Tab.~\ref{tab:Localization}, our method achieves an overall accuracy of 87.0\%, about 10 points higher than LangSplat and 15\% better than LERF. Furthermore, our method achieves the highest localization accuracy in three out of four scenes, demonstrating its superiority in open-vocabulary retrieval abilities.

% Our method enables highly fine-grained localization, such as pinpointing Pikachu's ears, hat, and cloak, as illustrated in the Fig.~\ref{fig:pika}. In contrast, approaches that segment and then input to CLIP fail to achieve this level of detail. These methods lose global semantic information; for instance, if Pikachu's ear is cropped out, it becomes a small black pixelated area that CLIP cannot recognize.

\begin{figure}[htb]
  \begin{minipage}[t]{1\linewidth} % 将图像包裹在 minipage 环境中，使其宽度为 0.6 行宽
    \includegraphics[width=\linewidth]{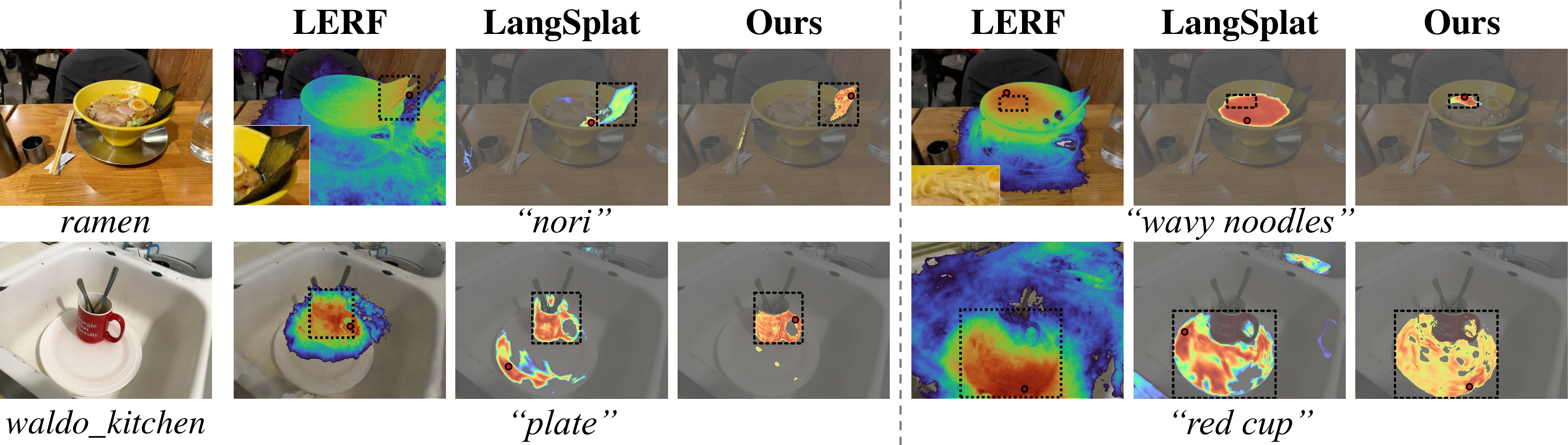}
    \caption{Comparison of the generated relevance score maps on the LERF dataset: the black bounding boxes are the ground truth and the red points are the localization results. Notice, the ground-truth of `wavy noodles' is only a small corner area and our result is very accurate.}
    \label{fig:loc_compare}
  \end{minipage}
\end{figure}

We also compare the retrieval performances qualitively and show the results in Fig.~\ref{fig:sofa_com}.
We visualize the relevancy scores following LERF into a 2D map. Notice that, LERF and LangSplat necessity computation of the relevancy score across multiple semantic levels, while our method only requires one time computation and enables fine-grained localization at the same time. As illustrated in Fig.~\ref{fig:loc_compare}, the highest score positions of LangSplat are wrong \clearpage for `nori', `wavy noodles' and the `red cup', whereas our method accurately locates them. Though the location of the reported highest score is correct for LERF, the activated regions of LERF are very scattered. In contrast, our method produces very clear boundaries and more concentrated activation regions compared to those of LERF and LangSplat.

\begin{figure}[htb]
  \begin{minipage}[t]{1\linewidth} % 将图像包裹在 minipage 环境中，使其宽度为 0.6 行宽
  \vspace{-4mm}
    \includegraphics[width=\linewidth]{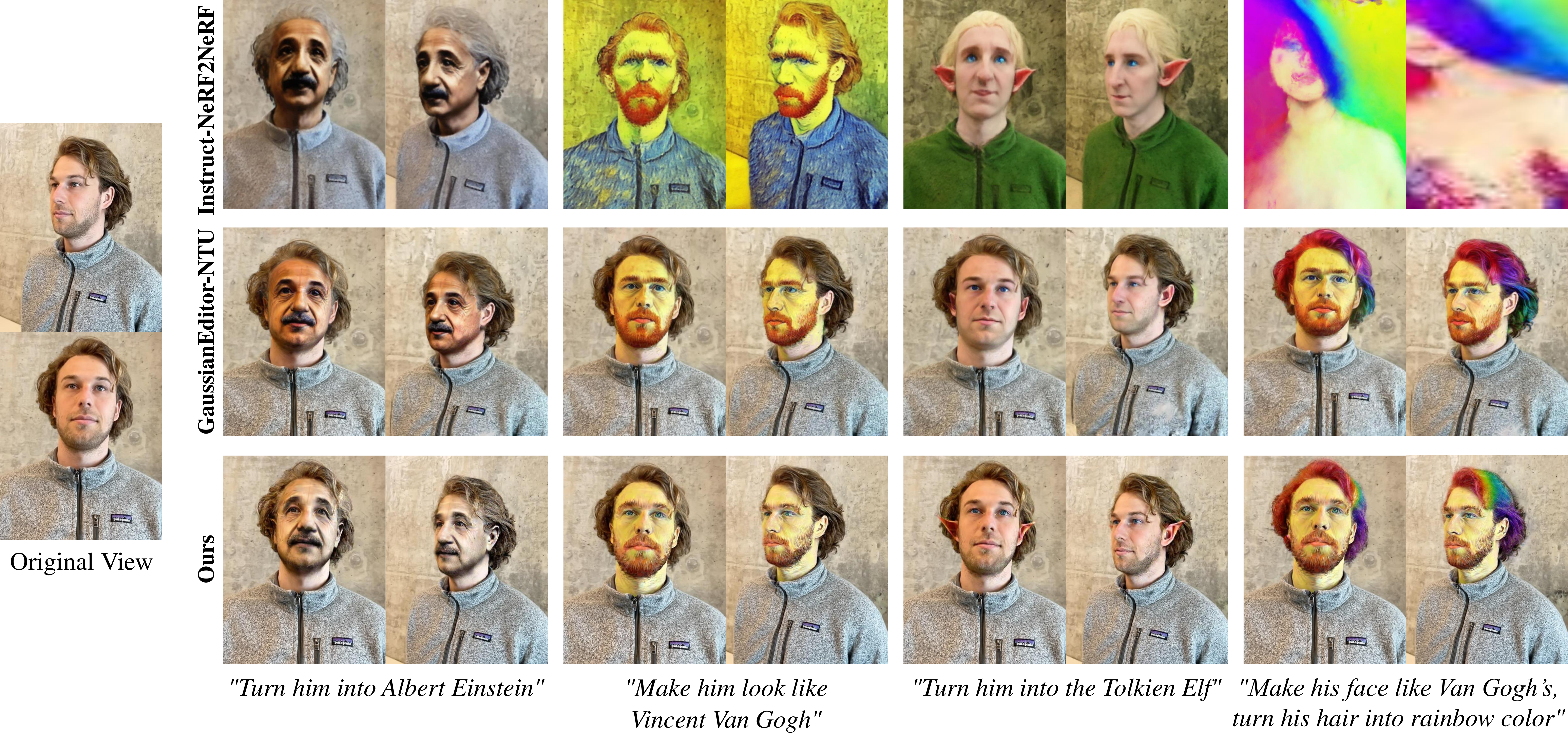}
    \caption{Qualitative comparison: Our method enables various types of portrait editing, including celebrities, artistic styles, characters from fantasy novels, and well support composite edits. Our edits exhibit very fine details, realistic and are view-consistent.}
    \label{fig:face}
  \end{minipage}
\end{figure}

\begin{figure}[htb]
  \begin{minipage}[t]{1\linewidth}
  \vspace{-4mm}
    \includegraphics[width=\linewidth]{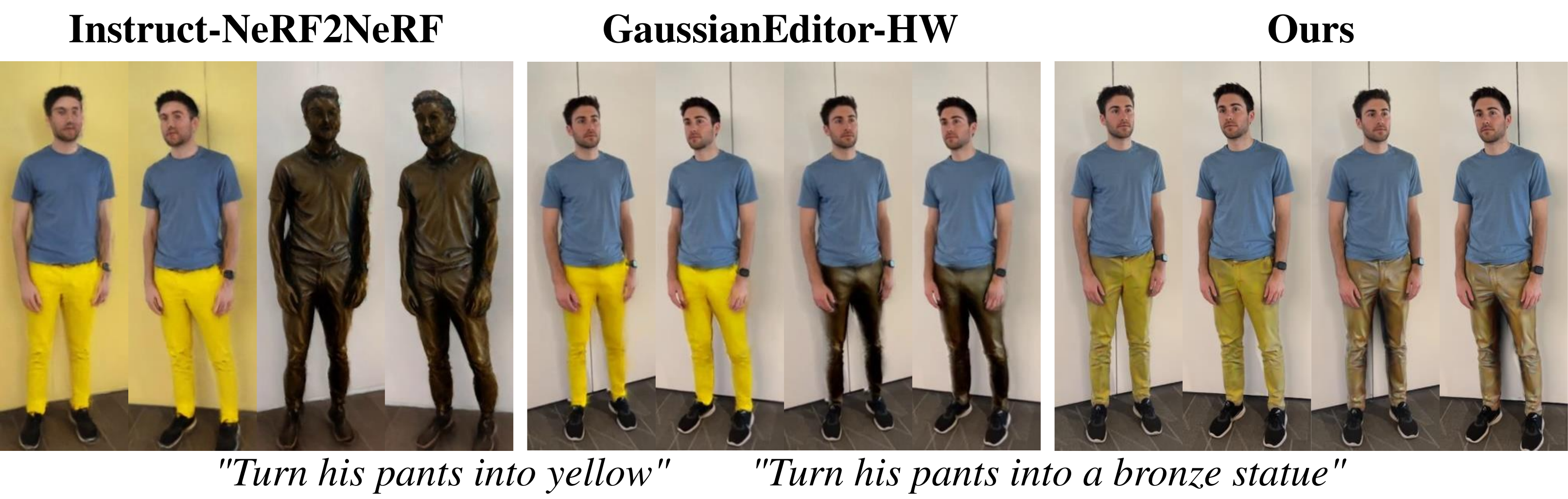}
    \caption{Qualitative comparison: Instruct-NeRF2NeRF is unable to perform partial edits, and GaussianEditor-HW produces blurry and inconsistent results. In contrast, our method generates precise and multi-view consistent human part edits.}
    \label{fig:person_compare}
    \vspace{-2mm}
  \end{minipage}
\end{figure}

\begin{figure}[htb]
  \begin{minipage}[t]{1\linewidth} % 将图像包裹在 minipage 环境中，使其宽度为 0.6 行宽
  \vspace{-4mm}
    \includegraphics[width=\linewidth]{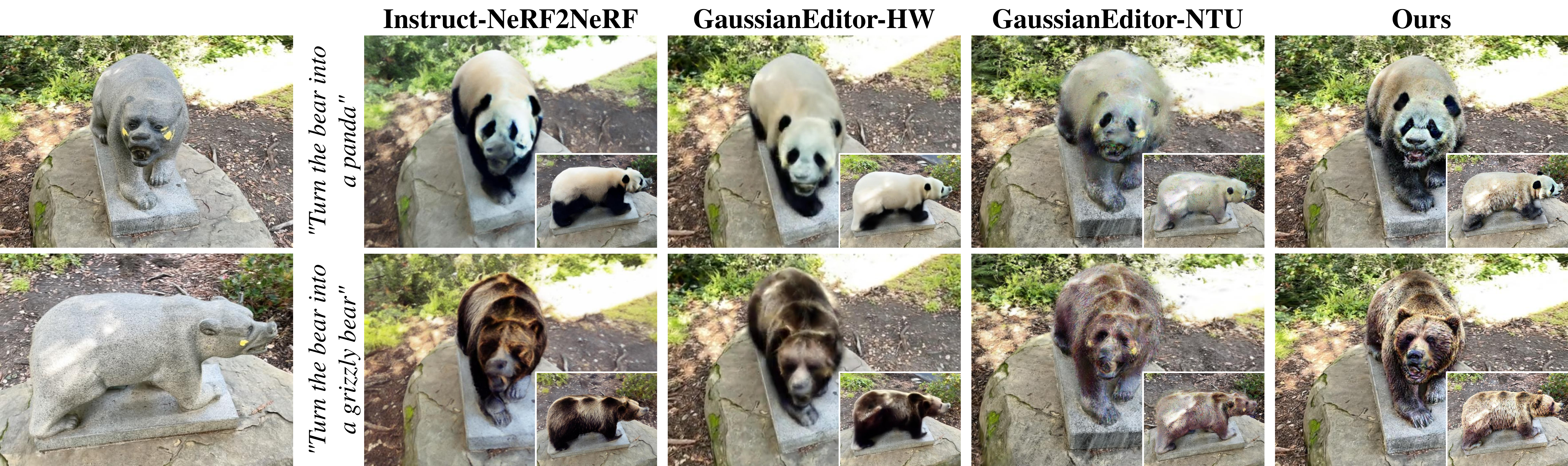}
    \caption{Qualitative comparison: our results of `panda' and `grizzly bear' show very fine fur and facial details. While other approaches suffer over-smoothing and the multi-face Janus problem.}
    \label{fig:bear}
    \vspace{-4mm}
  \end{minipage}
\end{figure}

% \vspace{-20mm}
\begin{figure}[htb]
  \begin{minipage}[t]{1\linewidth} % 将图像包裹在 minipage 环境中，使其宽度为 0.6 行宽
  % \vspace{-4mm}
    \includegraphics[width=\linewidth]{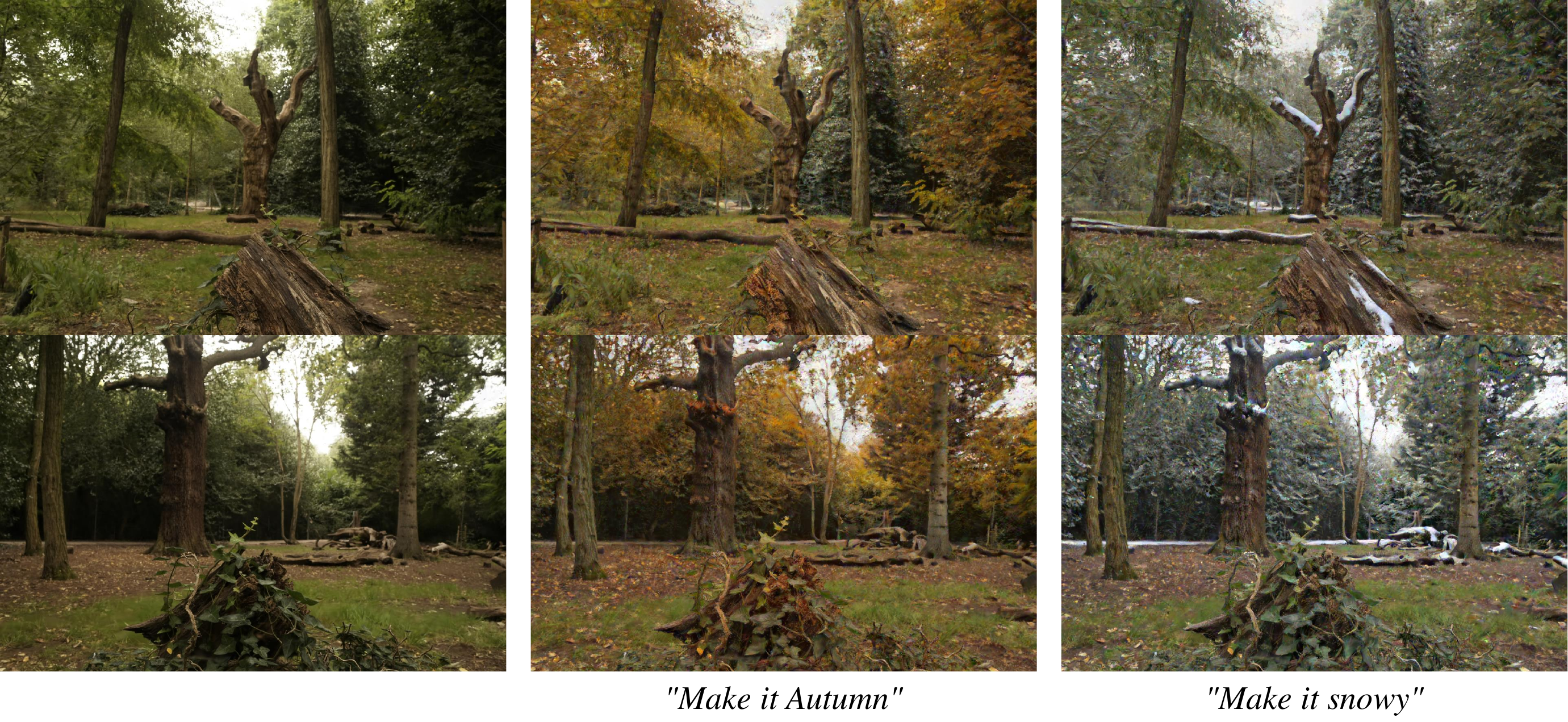}
    \caption{Qualitative results: environmental changes.}
    \label{fig:stump}
    \vspace{-6mm}
  \end{minipage}
\end{figure}

\begin{figure}[htb]
  \begin{minipage}[t]{1\linewidth} % 将图像包裹在 minipage 环境中，使其宽度为 0.6 行宽
  \vspace{-6mm}
    \includegraphics[width=\linewidth]{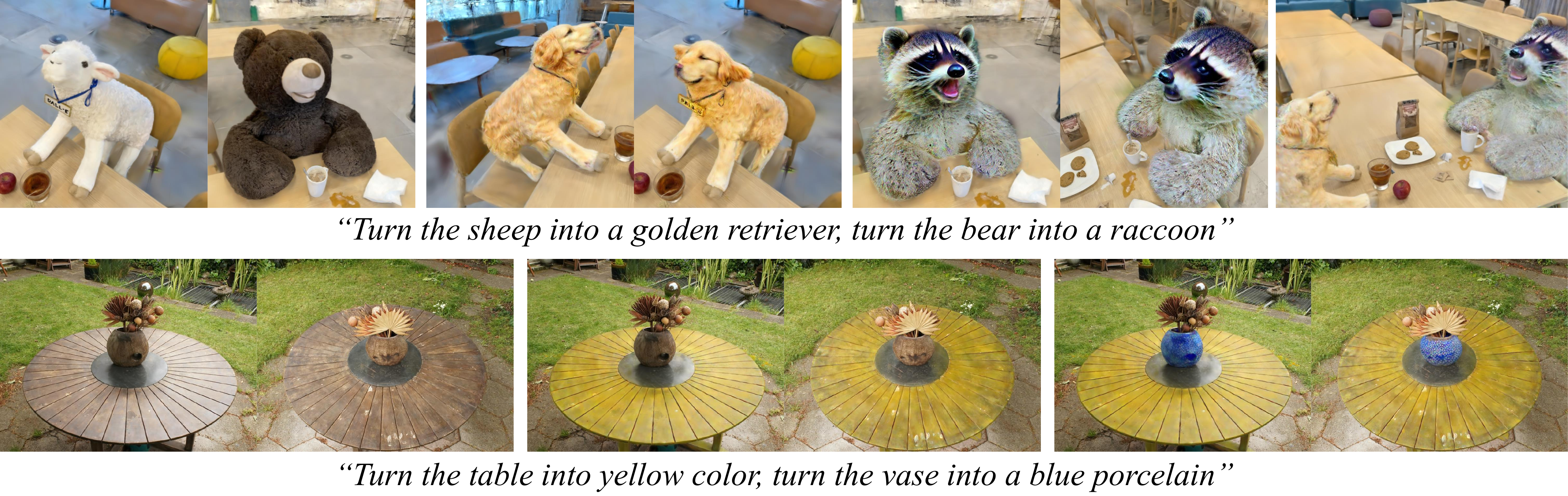}
    \caption{Compositional editing: Our approach enables multi-object compositional editing without mutual interference. The edits show very fine details. Notice the blurriness in background of the `tea-time' scene is due to input blurry images in the dataset.}
    \label{fig:2_step}
  \end{minipage}
\end{figure}

\textbf{3D Gaussian Editing}
We compare our 3D Gaussian editing performance with Instruct-NeRF2NeRF~\cite{haque2023instruct}, GaussianEditor-NTU~\cite{chen2023gaussianeditor} and GaussianEditor-HW~\cite{fang2023gaussianeditor} on various datesets including the 'bear' and 'face' scene from Instruct-NeRF2NeRF. We implement our language embedding on LangSplat~\cite{qin2023langsplat} and editing for 3D Gaussian on Threestudio~\cite{threestudio2023}. All our experiments were conducted on a single Tesla V100S GPU. Depending on the prompt and the complexity of the scene, the editing process typically requires optimization for 1200 to 3000 steps, taking approximately 10 to 30 minutes in total. More experimental details will be provided in the appendix. The experiments results are shown in Fig.~\ref{fig:bear} and Fig.~\ref{fig:face} repsectively. As we can see, for the bear scene in Fig.~\ref{fig:bear}, our results contain much finer details and the bear heads of the edited results are consistent. In contrast, other approches suffer over-smoothing and Janus problems to different extent. Moreover, for the 'face' scene, our eidts are more consistent with the original video and more coherent. Instruct-NeRF2NeRF doens't support multi-round editing, while GaussianEditor-NTU generates blurry faces, and fails for the `\textit{Tolkien Elf}' edits.

%evaluate our method on LERF, Instruct-NeRF2NeRF and Mip-NeRF360 dataset. We utilize the renderer from 3D Gaussian~\cite{kerbl20233d} for 

\subsection{Qualitative Evaluation}
% \begin{wrapfigure}{r}{0.5\linewidth} % 'r' 表示图像靠右，宽度为 0.7 行宽
%   \centering
%   \includegraphics[width=\linewidth]{fig/pika.pdf}
%   \caption{Fine-grained localization.}
%   \label{fig:pika}
% \end{wrapfigure}

To demonstrate that our TIGER can handle various scenes and prompts, we visualize more results of our TIGER on environment editing prompts and compositional editing prompts as shown in Fig.~\ref{fig:stump} and Fig.~\ref{fig:2_step} respectively.
As show in Fig.~\ref{fig:stump}, our TIGER edits the 3D Gaussian scene in very subtle way, see the yellow leaves for `autumn' and snow in `snowy' edits. Moreover, we our method can handle large appearance edits as shown in Fig.~\ref{fig:2_step}. The results of `dog' and `raccoon' are very accurate and realistic, and the `table' and `vase' also show very subtle change, while preserving the textures of original object.
%Our qualitative results are illustrated in the Fig.~\ref{fig:face}, Fig.~\ref{fig:bear} and Fig.~\ref{fig:2_step}.

%Benefiting from relevance scores, we are able to achieve precise object-level editing and composite editing, as depicted. It can be observed that our edits are finer compared to previous methods, such as human eyes, hair, and bear fur. Moreover, larger geometric transformations can be achieved, such as the bear's head and the elf's ears. 
%Additionally, we can also realize edits on the overall scene environment, as illustrated in the Fig.~\ref{fig:stump}. The results of GaussianEditor and Instruct-NeRF2NeRF were generated using the default configurations from their respective official repositories.

\subsection{Ablations}

To demonstrate our design choices, we conduct ablation studies on using MVDream for CSD, and our score-based 3D Guassian updating.

\begin{figure}[htb]
  \begin{minipage}[t]{0.5\linewidth}
    \centering
    \includegraphics[width=\linewidth]{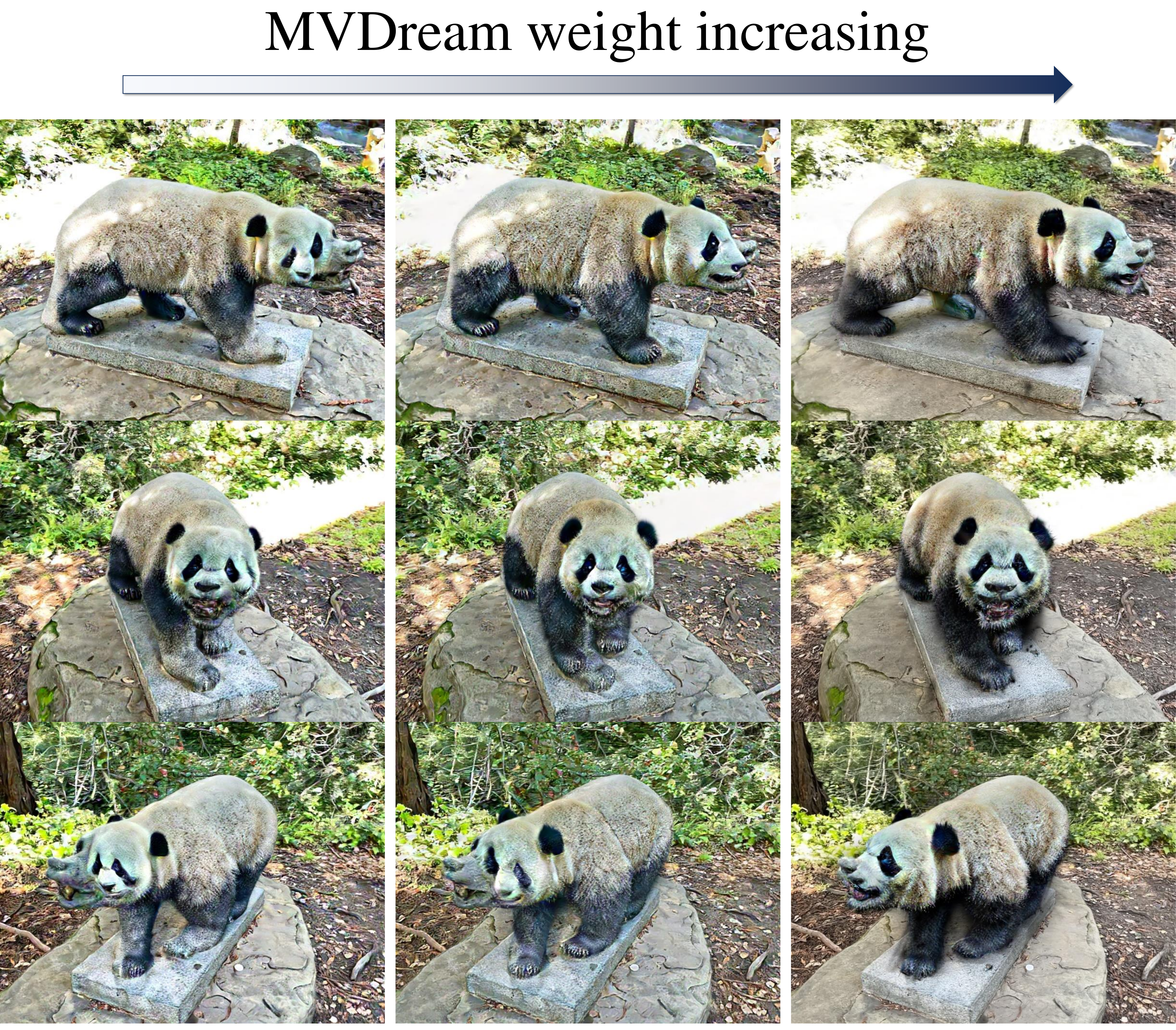}
    \caption{Ablation: MVDream alleviate the multi-face Janus issue.}
    \label{fig:ab_mv_1}
  \end{minipage}
  \hfill
  \begin{minipage}[t]{0.45\linewidth}
    \centering
    \includegraphics[width=\linewidth]{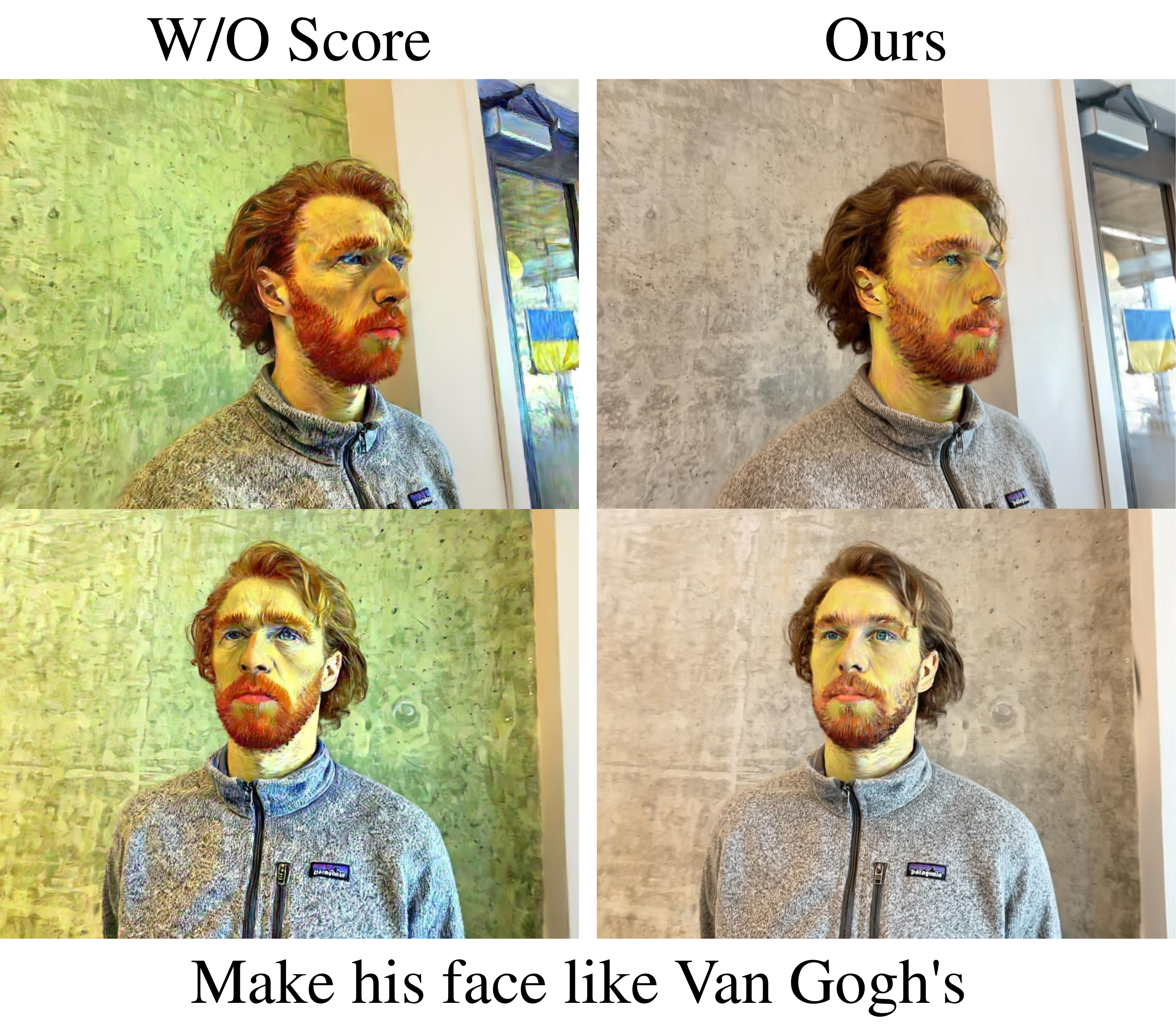}
    \caption{Ablation: score-base updating limits the editing area.}
    \label{fig:ab_mv_2}
  \end{minipage}
\end{figure}

\textbf{MVDream.}
We demonstrate effectiveness of MVDream in CSD by gradually increasing the MVDream loss weight and show result in Fig.~\ref{fig:ab_mv_1}. It can be seen that as the MVDream weight increases, the multi-face Janus issue gradually alleviated.

\textbf{Relevance score based updating.}
To demonstrate the effect of score constraints on Gaussian updates, we demonstrate the results without the constraint of relevance score on the gradient in Fig.~\ref{fig:ab_mv_2}. As we can see that, the modifications tend to affect the entire scene, background wall turns yellowish, without score based updating constraints.

% \begin{figure}[htb]
%   \begin{minipage}[t]{0.5\linewidth} % 将图像包裹在 minipage 环境中，使其宽度为 0.6 行宽
%     \includegraphics[width=\linewidth]{fig/ab_mv.pdf}
%     \caption{Ablation.}
%     \label{fig:ab_mv}
%   \end{minipage}
% \end{figure}
\section{Conclusion}
\label{sec:con}
This paper proposes a systematic approach, namely TIGER, for coherent text-instructed 3D Gaussian retrieval and editing.TIGER adopts a bottom-up language aggregation strategy to generate a denser language embedded 3D Gaussians that supports open-vocabulary retrieval. It also incorporates a Coherent Score Distillation (CSD) that aggregates a 2D image editing diffusion model and a multi-view diffusion model, producing multi-view consistent editings with much more fine details. In various experiments, our TIGER is able to accomplish more consistent and realistic edits than prior work.

\paragraph{Limitations.}
We adopt the MaskCLIP to extract language features, hence it also suffers the ``bag-of-words'' problem, where phrases like ``not red'' are treated similarly to ``red''. 
Our editing process depends on pre-trained 2D diffusion models, it is limited in handling highly complex instructions. Additionally, the score distillation process requires up-to 30 minutes for the most extensive edits. Although it's comparable to, and occasionally quicker than, GaussianEditor-NTU based on our testing, such prolonged processing times are still deemed excessively lengthy for practical user applications. %dataset udpate approaches (7 mins for GaussianEditor-NTU). 

{
    \small
    \bibliographystyle{nips}
    \bibliography{main}
}
\newpage
\appendix

\section{Appendix}
\label{sec:App}
\subsection{Additional implementation details}
In order to maintain consistency across multiple viewpoints, view dependent is cancelled. The Classifier-free guidance weights for InstructPix2Pix follow the default settings, with $S_I=1.5$ and $S_T=7.5$. For MVDream, the parameters from the official 3D generation repository are used, except for reducing max\_step\_percent to 0.2 to ensure update stability. The initial weight ratio of InstructPix2Pix to MVDream is set at $2:1$. Over time, the weight for InstructPix2Pix gradually increases while the weight for MVDream decreases until MVDream's weight reaches zero after 75\% of the total training epochs. For scenes with environmental and minimal viewpoint changes, editing can rely entirely on InstructPix2Pix. FeatUp is implemented using the implicit upsampler from its official repository, with parameters following the default settings.

\begin{wrapfigure}{r}{0.5\linewidth} % 'r' 表示图像靠右，宽度为 0.7 行宽
  \centering
  \vspace{-5mm} % 负值表示缩小垂直距离
  \includegraphics[width=\linewidth]{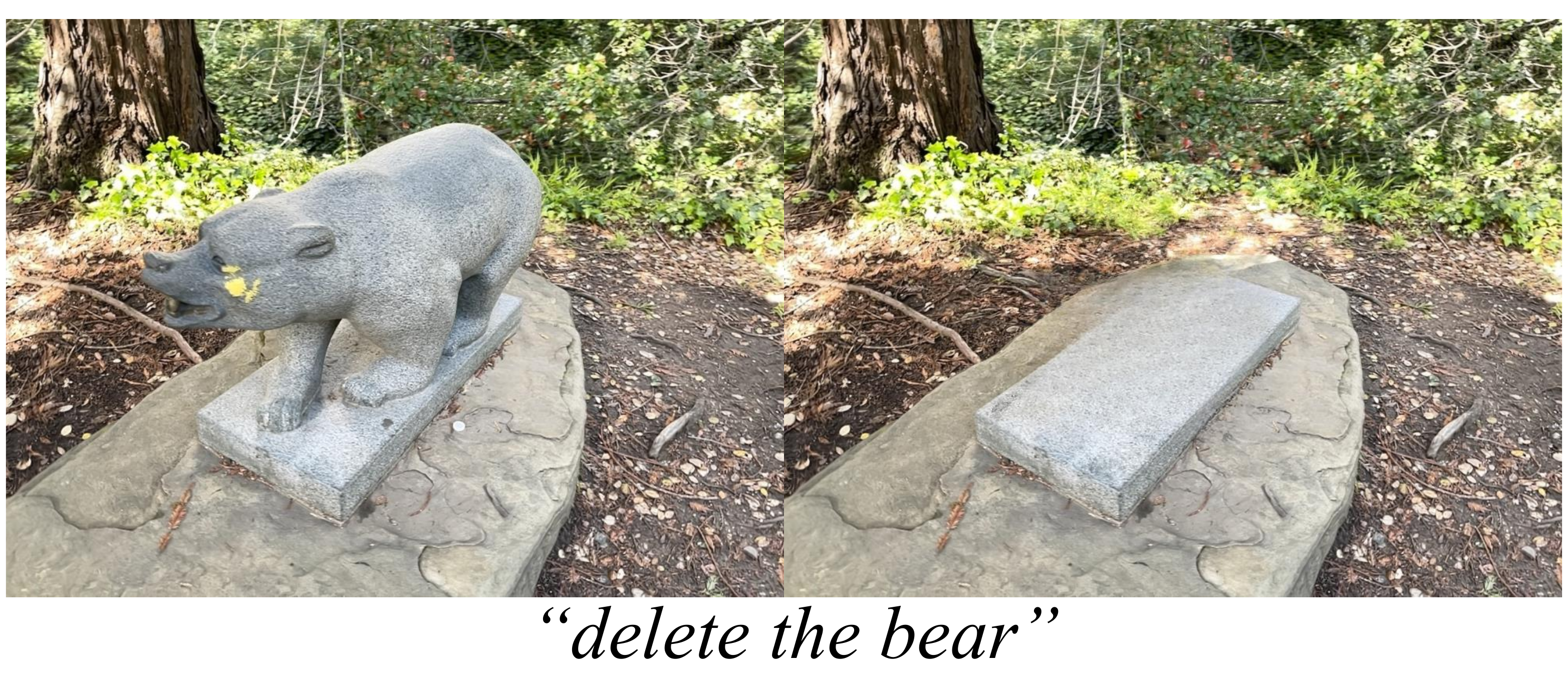}
  \caption{Delete the bear.}
  \label{fig:delete_bear}
\end{wrapfigure}

\subsection{Delete objects}
% \begin{figure}[htb]
%   \begin{minipage}[t]{0.5\linewidth} % 将图像包裹在 minipage 环境中，使其宽度为 0.6 行宽
%     \includegraphics[width=\linewidth]{fig/delete_bear.pdf}
%     \caption{delete the bear.}
%     \label{fig:delete_bear}
%   \end{minipage}
% \end{figure}

We eliminate the queried object from 3D Gaussians, render the remainings into training views, and identify holes from alpha map, which are inpainted using LaMa. We use inpainted images to retrain the 3D Gaussians. While texture consistency varies, this approach is effective for simple geometries, creating visually appealing results like the stone platform in Fig.~\ref{fig:delete_bear}.

\subsection{More results}

\begin{figure}[htb]
  \begin{minipage}[t]{1\linewidth} % 将图像包裹在 minipage 环境中，使其宽度为 0.6 行宽
    \includegraphics[width=\linewidth]{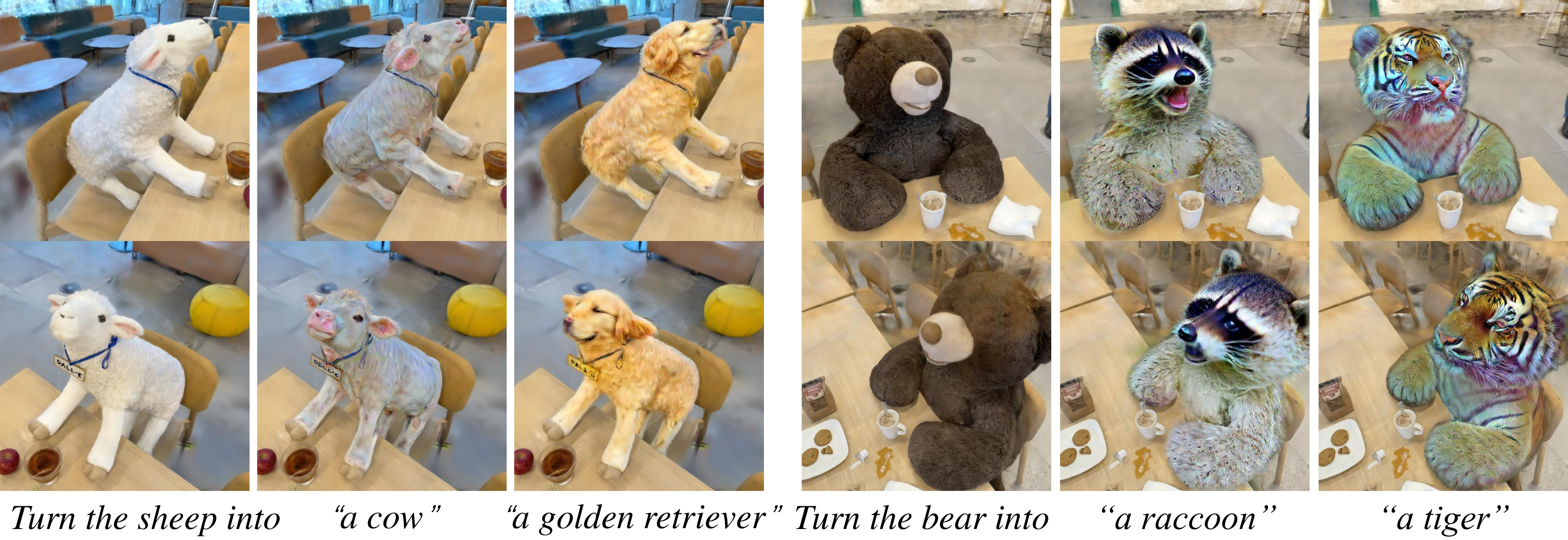}
    \caption{More results of TIGER on the `tea time' scene.}
    \label{fig:teatime}
  \end{minipage}
\end{figure}

\begin{figure}[htb]
  \begin{minipage}[t]{1\linewidth} % 将图像包裹在 minipage 环境中，使其宽度为 0.6 行宽
    \includegraphics[width=\linewidth]{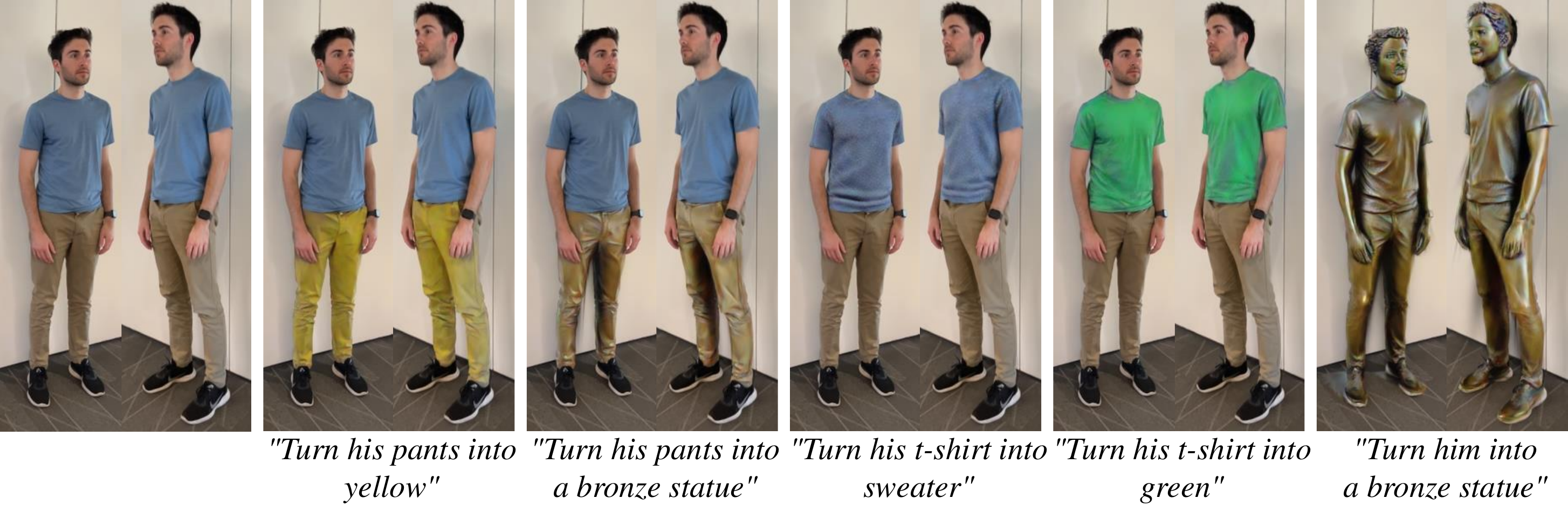}
    \caption{More results of TIGER on partial body editing.}
    \label{fig:person}
  \end{minipage}
\end{figure}

% \begin{figure}[htb]
%   \begin{minipage}[t]{1\linewidth} % 将图像包裹在 minipage 环境中，使其宽度为 0.6 行宽
%     \includegraphics[width=\linewidth]{fig/person_compare.pdf}
%     \caption{More results of TIGER on partial body editing.}
%     \label{fig:person_compare}
%   \end{minipage}
% \end{figure}

\begin{figure}[htb]
  \begin{minipage}[t]{1\linewidth} % 将图像包裹在 minipage 环境中，使其宽度为 0.6 行宽
    \includegraphics[width=\linewidth]{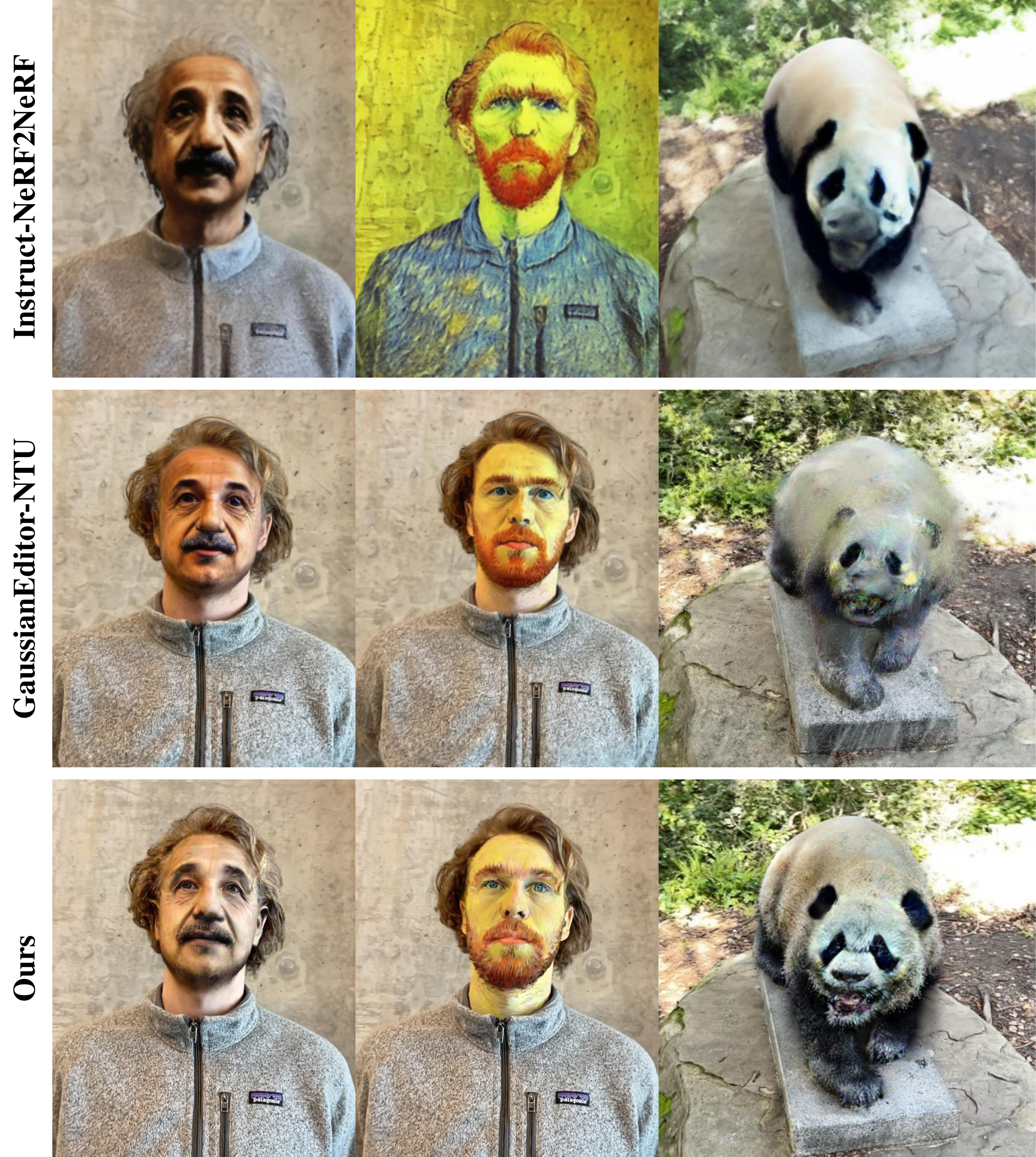}
    \caption{Details comparison.}
    \label{fig:detail}
  \end{minipage}
\end{figure}
\begin{figure}[htb]
  \begin{minipage}[t]{1\linewidth} % 将图像包裹在 minipage 环境中，使其宽度为 0.6 行宽
    \includegraphics[width=\linewidth]{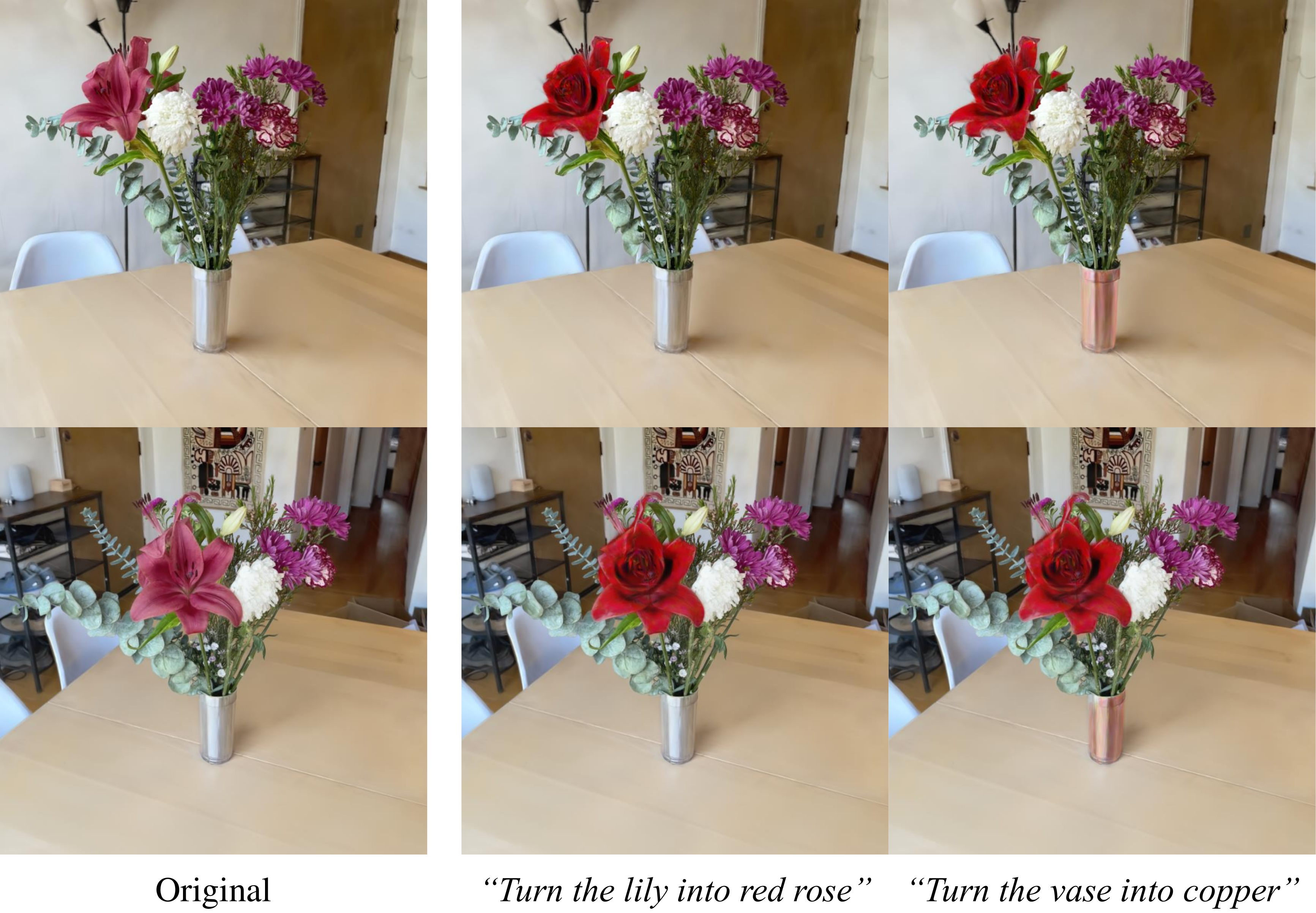}
    \caption{Details.}
    \label{fig:teaser_detail}
  \end{minipage}
\end{figure}
\begin{figure}[htb]
  \begin{minipage}[t]{1\linewidth} % 将图像包裹在 minipage 环境中，使其宽度为 0.6 行宽
    \includegraphics[width=\linewidth]{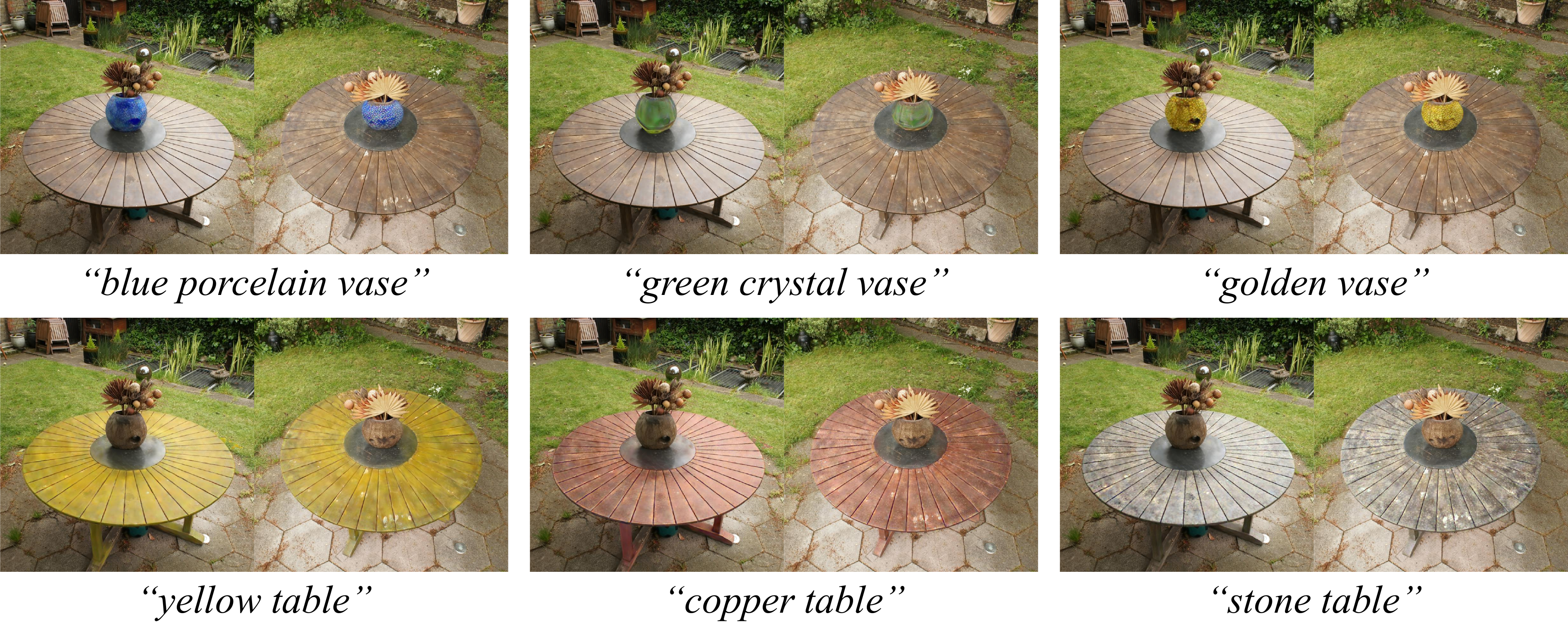}
    \caption{More results of TIGER.}
    \label{fig:garden}
  \end{minipage}
\end{figure}

\end{document}